\documentclass[a4paper,fleqn]{cas-dc}

\usepackage[numbers]{natbib}

\usepackage{amsmath,amsfonts}
\usepackage{algorithmic}
\usepackage{algorithm}
\usepackage{array}
\usepackage{booktabs}
\usepackage[caption=false,font=normalsize,labelfont=sf,textfont=sf]{subfig}
\usepackage{textcomp}
\usepackage{stfloats}
\usepackage{url}
\usepackage{verbatim}
\usepackage{cleveref}
\usepackage{graphicx}
\usepackage{multirow}
\usepackage{xcolor}

\let\oldenumerate\enumerate
\renewcommand{\enumerate}{
  \oldenumerate
  \setlength{\itemsep}{3pt}
  \setlength{\parskip}{0pt}
  \setlength{\parsep}{3pt}
}

\begin{document}

\title [mode = title]{Small Target Detection Based on Mask-Enhanced Attention Fusion of Visible and Infrared Remote Sensing Images}         

\author[1,2,3]{Qianqian Zhang}\cormark[2]
\ead{zhangqianqian21@mails.ucas.ac.cn }
\author[3]{Xiaolong Jia}
\author[3]{Ahmed M. Abdelmoniem}
\author[1]{Li Zhou}\cormark[1]
\ead{zhouli@nssc.ac.cn}

\author[1,4]{Junshe An}

\cortext[cor1]{Corresponding author}
\cortext[cor2]{Principal corresponding author}

\affiliation[1]{organization={National Space Science Center, Chinese Academy of Sciences}, city={Beijing}, postcode={100190}, country={China}}
\affiliation[2]{organization={School of Computer Science and Technology, University of Chinese Academy of Sciences}, city={Beijing}, country={China}}
\affiliation[3]{organization={School of Electronic Engineering and Computer Science, Queen Mary University of London}, city={London}, country={UK}}
\affiliation[4]{organization={School of Astronomy and Space Science, University of Chinese Academy of Sciences}, city={Beijing}, country={China}}

\begin{abstract}
Targets in remote sensing images are usually small, weakly textured, and easily disturbed by complex backgrounds, challenging high-precision detection with general algorithms. Building on our earlier ESM-YOLO, this work presents ESM-YOLO+ as a lightweight visible infrared fusion network. To enhance detection, ESM-YOLO+ includes two key innovations. (1) A Mask-Enhanced Attention Fusion (MEAF) module fuses features at the pixel level via learnable spatial masks and spatial attention, effectively aligning RGB and infrared features, enhancing small-target representation, and alleviating cross-modal misalignment and scale heterogeneity. (2) Training-time Structural Representation (SR) enhancement provides auxiliary supervision to preserve fine-grained spatial structures during training, boosting feature discriminability without extra inference cost. Extensive experiments on the VEDAI and DroneVehicle datasets validate ESM-YOLO+'s superiority. The model achieves 84.71\% mAP on VEDAI and 74.0\% mAP on DroneVehicle, while greatly reducing model complexity, with 93.6\% fewer parameters and 68.0\% lower GFLOPs than the baseline. These results confirm that ESM-YOLO+ integrates strong performance with practicality for real-time deployment, providing an effective solution for high-performance small-target detection in complex remote sensing scenes.
\end{abstract}


\begin{keywords}
Multimodal fusion\sep Visible\sep Infrared\sep Small target\sep Detection 

\end{keywords}

\maketitle

\section{Introduction}

As artificial intelligence advances, target detection techniques on satellites and unmanned aerial vehicles (UAVs) have become pivotal for civilian~\cite{10462171, 10335666} and military~\cite{10282104, 10281409} target detection. Despite their utility, UAV-based remote-sensing imagery poses significant challenges for detecting small targets. The limited pixel information from small, densely distributed targets, combined with complex backgrounds, hinders effective feature extraction and representation, ultimately reducing detection accuracy. Furthermore, because visible-light imaging relies on external illumination, variations in lighting caused by weather conditions further complicate target detection in remote-sensing imagery.


In contrast to visible-light imaging, infrared imaging relies on passive thermal radiation sensing, offering inherent robustness to illumination variations and environmental interference, thereby facilitating more stable perception of small targets under challenging conditions. However, the absence of fine-grained shape and texture details in infrared imagery limits its discriminative power for accurate detection of small targets. To address this limitation, multispectral visible–infrared fusion exploits the complementary strengths of RGB structural and texture information and infrared thermal saliency to enhance feature representation. For example, Tao Tian et al.~\cite{10282173} and Jiaqing Zhang et al.~\cite{10075555} demonstrated that such cross-modal integration significantly improves small target recognition performance.

Despite the complementary nature of visible and infrared modalities, multimodal small-target detection remains a challenge. First, cross-modal feature heterogeneity in scale, texture, and thermal signatures complicates discriminative feature extraction. Second, spatial and temporal misalignment due to sensor viewpoints, acquisition timing, or UAV motion degrades fusion quality for low-saliency targets. Finally, practical deployment constraints, including limited annotated datasets, high computational cost, and real-time requirements, limit the generalization of algorithms.

Existing multimodal small-target detectors predominantly emphasize feature enhancement through increasingly complex fusion designs. For instance, MIR-YOLO~\cite{zhong2025mir} leverages gated aggregation and Vision-LSTM to strengthen contextual modeling, while DVIF-Net~\cite{zhao2025dvif} performs dual-branch extraction with channel attention for cross-modal enhancement. Although effective, these approaches treat fusion primarily as a representational enrichment problem, typically introducing heavy architectures that are unsuitable for resource-constrained UAV and satellite platforms. 
More importantly, they either assume perfectly aligned modalities or depend on intensive feature encoding, leaving cross-modal structural inconsistency largely unaddressed.


We argue that the fundamental challenge of visible-infrared small-target detection lies not in aggressive feature enhancement, but in constructing a modality-invariant and spatially consistent representation under heterogeneous sensing statistics and imperfect spatial correspondence. Direct fusion implicitly performs unconditional aggregation, which may amplify modality-dominant background responses or suppress spatially localized weak targets, resulting in representations that are neither statistically comparable nor geometrically stable across scales.

Effective fusion should encode a spatial reliability prior that conditions cross-modal interaction on trustworthy regions while preserving fine-grained object support. This motivates a principled mask-then-attend formulation. Specifically, the proposed Mask-Enhanced Attention Fusion (MEAF) module decomposes fusion into two complementary operators: (i) a learnable spatial mask that performs soft alignment-aware selection to suppress unreliable interactions, and (ii) spatial attention that conducts topology-consistent reweighting to maintain small-object spatial support after fusion. By transforming fusion from unconditional aggregation to reliability-conditioned interaction, MEAF directly addresses cross-modal heterogeneity and scale misalignment at the structural level.

To this end, we propose ESM-YOLO+, a lightweight visible–infrared fusion network with targeted architectural enhancements. It introduces two modules: (i) Mask-Enhanced Attention Fusion (MEAF) module that performs pixel-level fusion using learnable spatial masks and spatial attention, improving small target representation and cross-modal scale handling in a lightweight, single-branch design; and (ii) training-time Structural Representation (SR) enhancement that provides gradient-level supervision to preserve fine-grained spatial structures, applied only during training to avoid any inference overhead. Together, these mechanisms enable robust multimodal feature extraction at low computational cost while improving small-target detection in complex remote sensing scenarios, as validated on the VEDAI~\cite{Razakarivony2016Vehicle} and DroneVehicle~\cite{9759286} datasets.

Our contributions are summarized as follows.
\begin{itemize}
    \item We propose ESM-YOLO+, a lightweight visible-infrared fusion network for real-time detection of small targets in remote sensing imagery, outperforming the baseline ESM-YOLO in both accuracy and efficiency.
    \item The Mask-Enhanced Attention Fusion (MEAF) module achieves pixel-level fusion via learnable spatial masks, effectively aligning RGB and IR features, enhancing the representation of small targets, and mitigating cross-modal misalignment and scale heterogeneity.
    \item Training-time Structural Representation (SR): provides auxiliary supervision to preserve fine-grained spatial structures during training, improving feature discriminability without increasing inference cost.
    \item Experiments on VEDAI and DroneVehicle datasets show ESM-YOLO+ improves VEDAI mAP by 2.29\%, reaches 74.0\% DroneVehicle accuracy, cuts 93.6\% parameters and 68.0\% GFLOPs, confirming robust performance and real-time practicality.
\end{itemize}


The rest of the paper is organized as follows. We first present the related work in Sec.~\ref{ch:RW}. Then we discuss ESM-YOLO and ESM-YOLO+’s key design in Sec.~\ref{ch:Me}. Our results are presented in Sec.~\ref{ch:Result}.  We discuss the implications of our design in Sec.~\ref{ch:discussion} and conclude the paper in Sec.~\ref{ch:Conclusion}.

\begin{figure*}[htbp]
\centering
\includegraphics[width=0.9\textwidth]{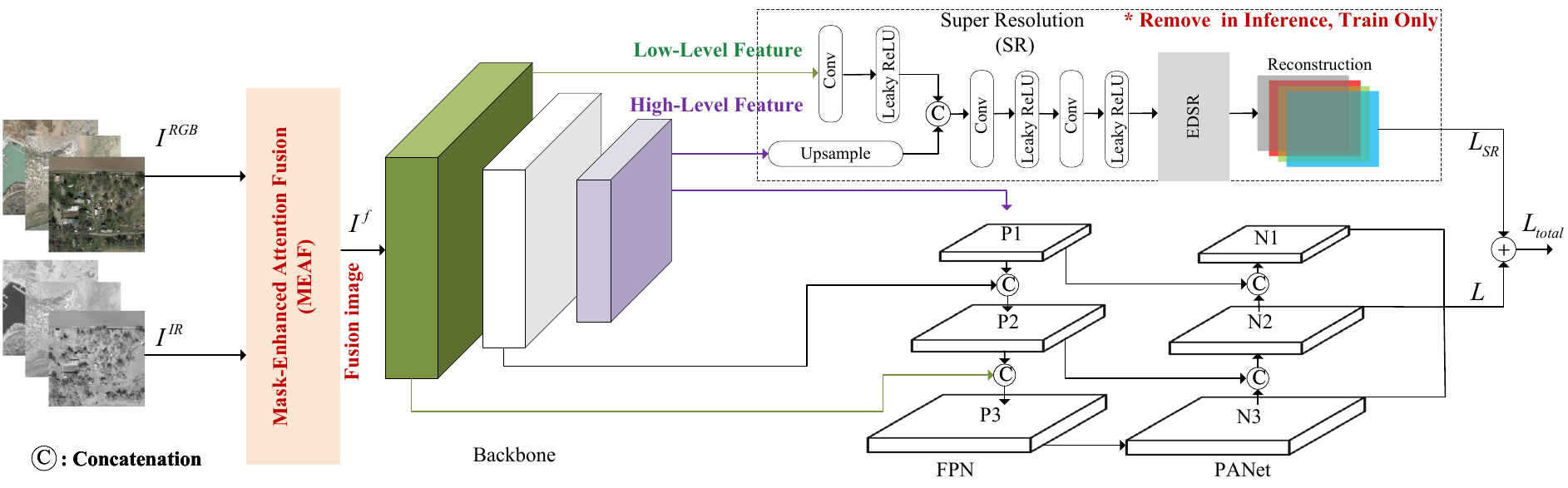}
\caption{{Overall architecture of the ESM-YOLO+. It comprises three key components: 1) the Mask-Enhanced Attention Fusion (MEAF) Module; 2) Detection Backbone and detection head; and 3) EDSR-based super-resolution branch used only during training to enhance spatial learning and removed at inference for faster detection.} }
\label{fig:1}
\end{figure*}

\section{Related Work}\label{ch:RW}
\subsection{Object Detection Using Multimodal Data}
Multimodal fusion integrates diverse sensor data to boost object detection in real-world scenarios, with applications in autonomous driving~\cite{9706418}, tunnel defect inspection~\cite{9708607}, and remote sensing classification~\cite{9955391}. By exploiting complementary RGB, IR, and SAR cues, multimodal methods often achieve higher accuracy and robustness than single-modal ones, especially in complex remote sensing environments.

Recent studies have validated the effectiveness of cross-modal integration in remote sensing. Infrared-visible fusion improves UAV fire detection~\cite{9953997} and reduces missed and false detections under varying illumination~\cite{10476333}, showing that multimodal fusion enhances adaptability and perceptual stability across complex conditions.

Multimodal fusion algorithms are generally classified into pixel-level~\cite{Liu2022Comparative}, feature-level~\cite{xin2022vehicle}, and decision-level fusion~\cite{Xiao2023Decision} according to their integration stage. Pixel-level fusion combines raw sensor data during preprocessing to retain rich original information, making it favorable for remote sensing. Feature-level fusion merges high-level representations, but can introduce feature loss and increase the risk of missed detections. Decision-level fusion aggregates outputs of different modalities, yet tends to cause redundant computation and double-counting. In remote sensing applications, subtle cross-modal differences are often weakened during hierarchical feature extraction, making pixel-level fusion more suitable for preserving fine-grained information across modalities. However, conventional pixel-level schemes are sensitive to background clutter, lack effective cross-modal alignment, and involve heavy computation for high-resolution images.These issues motivate lightweight pixel-level fusion methods that strengthen discriminative representation while ensuring real-time efficiency. 


\subsection{Small Target Detection}


Small target detection is a challenging and vital task in computer vision. Compared with conventional targets, small targets occupy few pixels and are easily obscured by complex backgrounds, complicating reliable feature extraction, accurate localization, and multi-scale representation.


To address these challenges, extensive improvements have been made to the backbone network to enhance the extraction of small-target features. Chen et al.~\cite{10471667} replaced C3 units with RepVGG blocks, Zhong et al.~\cite{10170867} introduced the OSA-C3 module for one-shot aggregation. While these enhancements improve detection accuracy, many involve local optimizations or increased model complexity, demand higher computational resources, and may overemphasize certain features, which can reduce generalization and require task-specific tuning.

Complementing backbone enhancements, neck and feature fusion strategies have been developed to integrate information across scales and layers. Low-level details such as edges and textures are fused with high-level semantic features to improve holistic representation. Notable examples include Softpool-based SPP modules \cite{9711197}, dilated convolution in SPP \cite{10261079}, residual SPP structures \cite{9886379}, and SPPF combined with channel-spatial attention mechanisms such as CBAM \cite{10422783}. These methods enhance feature representation and detection accuracy, but they often incur substantial computational overhead or introduce artifacts, such as the grid effect in dilated convolutions, which can compromise spatial coherence and hinder real-time deployment.


Overall, while current approaches have improved small target detection through backbone refinement and multi-scale feature fusion, challenges remain in efficiently preserving fine-grained spatial features and maintaining computational efficiency. These limitations directly motivate the design of ESM-YOLO+, a lightweight pixel-level fusion network that enhances small-target representation while enabling real-time deployment in remote sensing applications.

\section{Method}
\label{ch:Me}

This section presents ESM-YOLO+, an enhanced multimodal, real-time detector for small-target detection in remote-sensing imagery. ESM-YOLO+ extends the previous conference version, ESM-YOLO\cite{esm}, by introducing a more discriminative pixel-level fusion mechanism and a training-time super-resolution guidance strategy. These enhancements substantially improve spatial representation learning and robustness against complex backgrounds. We first briefly summarize the baseline ESM-YOLO\cite{esm} architecture, and then detail the proposed methodological advancements that constitute the core contributions of ESM-YOLO+.
\subsection{Baseline: ESM-YOLO}

ESM-YOLO\cite{esm} is a one-stage multimodal object detector designed for real-time small target recognition in remote sensing scenarios. The network consists of three principal components: a pixel-level multimodal fusion module, a convolutional backbone for hierarchical feature extraction, and a detection head with multi-scale predictions.

At the input stage, visible and infrared images are fused using the Bilateral Excitation Fusion (BEF) module, which applies channel-wise excitation to leverage complementary modalities. The fused features are subsequently processed by a convolutional backbone incorporating Improved Atrous Spatial Pyramid Pooling (IASPP) and Compact BottleneckCSP (CBCSP) blocks to enhance multi-scale context aggregation while maintaining computational efficiency. Finally, a feature pyramid-based detection head produces objectness, localization, and classification predictions at multiple resolutions.

Despite its competitive performance, ESM-YOLO relies on shallow convolutional fusion without explicit spatial modeling, rendering it vulnerable to background clutter and insensitive to weak small targets. These limitations motivate the enhanced design of ESM-YOLO+.

\subsection{ESM-YOLO+: Enhanced Method}

\subsubsection{Mask-Enhanced Attention Fusion Module}

The quality of pixel-level multimodal fusion plays a critical role in small target detection, particularly in remote sensing imagery with complex backgrounds and low target saliency. In ESM-YOLO\cite{esm}, the BEF module performs modality fusion through simple convolutional excitation, which is insufficient to selectively emphasize informative regions while suppressing background interference.

To address this issue, ESM-YOLO+ introduces the Mask-Enhanced Attention Fusion (MEAF) module. MEAF extends BEF by jointly incorporating learnable spatial masks and explicit spatial attention, enabling the network to adaptively highlight target-relevant regions and enhance fine-grained structural details.

\begin{figure*}[pos=t]
\centering
\includegraphics[width=0.85\textwidth]{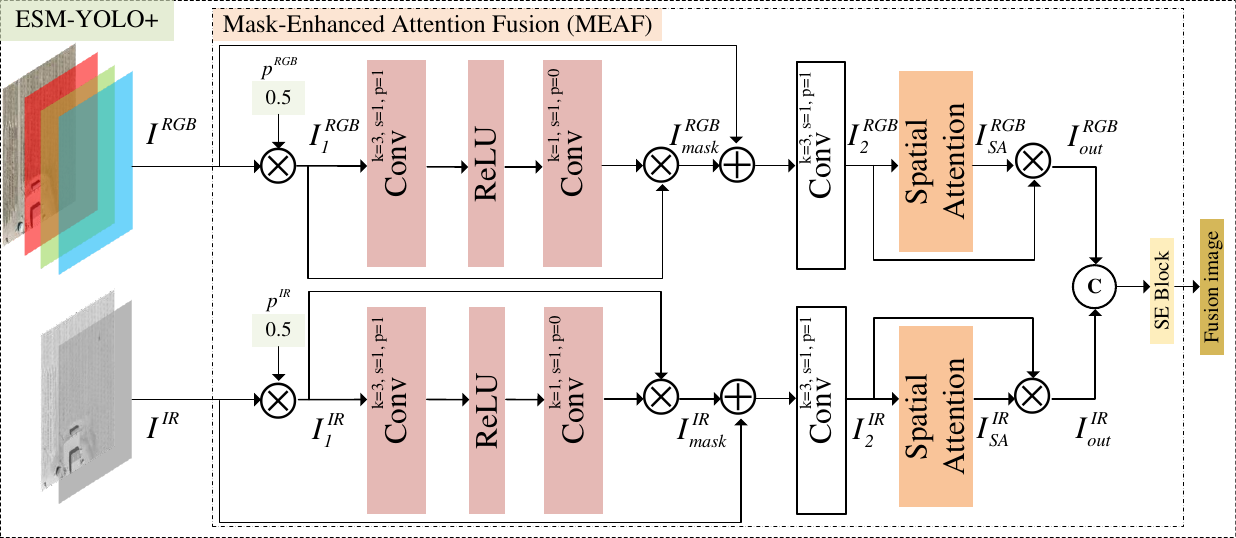}
\caption{Mask-Enhanced Attention Fusion (MEAF) Module. Pixel-level fusion module in the ESM-YOLO+ model.} 
\label{fig:8133}
\end{figure*}

Given an RGB image $I^{RGB} \in \mathbb{R}^{C \times H \times W}$ and an IR image $I^{IR} \in \mathbb{R}^{1 \times H \times W}$, modality-wise scaling is first applied:
\begin{equation}
\label{eq:1}
I^{RGB}_1 = I^{RGB} \cdot p^{RGB}, \quad
I^{IR}_1 = I^{IR} \cdot p^{IR},
\end{equation}
where $p^{RGB}$ and $p^{IR}$ are learnable modal parameters.

Then spatial masks are generated via convolutions to highlight salient structures:
\begin{equation}
\begin{aligned}
I^{RGB}_{mask} &= I^{RGB}_1 \otimes 
\text{Conv}_{1\times 1}\!\left(
\sigma_R\!\left(\text{Conv}_{3\times 3}(I^{RGB}_1)\right)
\right), \\
I^{IR}_{mask}  &= I^{IR}_1 \otimes 
\text{Conv}_{1\times 1}\!\left(
\sigma_R\!\left(\text{Conv}_{3\times 3}(I^{IR}_1)\right)
\right)
\end{aligned}
\end{equation}
where $\sigma_R$ denotes the ReLU activation.

The masked features are further refined through convolutional aggregation:
\begin{equation}
\begin{aligned}
I^{RGB}_2 &= \text{Conv}_{3\times 3}\!\left(I^{RGB}_{mask} + I^{RGB}\right), \\
I^{IR}_2  &= \text{Conv}_{3\times 3}\!\left(I^{IR}_{mask} + I^{IR}\right)
\end{aligned}
\end{equation}

To explicitly model spatial importance, a spatial attention mechanism is applied to each modality:
\begin{equation}
\begin{aligned}
I^{RGB}_{SA} &= \sigma_S\!\Big(
\text{Conv}_{1\times 1}
\big[
\text{Cat}\big(
\bar{I}^{RGB}_2,\,
\hat{I}^{RGB}_2
\big)
\big]
\Big), \\
I^{IR}_{SA}  &= \sigma_S\!\Big(
\text{Conv}_{1\times 1}
\big[
\text{Cat}\big(
\bar{I}^{IR}_2,\,
\hat{I}^{IR}_2
\big)
\big]
\Big)
\end{aligned}
\end{equation}
where $\bar{I}$ and $\hat{I}$ represent the channel-wise average-pooled and max-pooled feature maps, respectively, and $\sigma_S$ denotes the Sigmoid function.

The attended features are obtained by element-wise modulation:
\begin{equation}
I^{RGB}_{out} = I^{RGB}_2 \otimes I^{RGB}_{SA}, \quad
I^{IR}_{out}  = I^{IR}_2 \otimes I^{IR}_{SA}
\end{equation}

Finally, the modality-specific features are concatenated and reweighted by a channel-wise excitation vector $M$, yielding the fused representation:
\begin{equation}
I^{F} = M \cdot (\text{Cat}[I^{RGB}_{out}, I^{IR}_{out}])
\end{equation}
where $M$ is computed via global average pooling followed by fully connected transformations.

By jointly leveraging mask-based enhancement and spatial attention, MEAF significantly improves the discriminability of fused features, particularly for small targets embedded in cluttered backgrounds.


\subsubsection{Training-Time Structural Representation Enhancement}

In resource-constrained real-time detection scenarios, improving small-object representation typically requires architectural expansion or multi-scale refinement mechanisms, which inevitably increase inference latency. However, such modifications are often infeasible for edge deployment and real-time applications. To address this fundamental trade-off between representational capacity and inference efficiency, ESM-YOLO+ introduces a training-time structural representation(SR) enhancement paradigm. The core principle is to strengthen spatial discriminability during training while preserving the original inference pathway and computational complexity.

Specifically, the proposed SR paradigm is instantiated by attaching an auxiliary reconstruction pathway to intermediate stages of the backbone network during training. The selected backbone feature maps are treated as structural embeddings and fed into a lightweight decoder composed of progressive upsampling and convolutional layers. This pathway is to enforce structural consistency between backbone representations and a spatially aligned reference. In this way, the auxiliary reconstruction objective functions as a geometry-aware regularization mechanism, constraining feature embeddings to retain fine-grained spatial topology that is otherwise weakened by aggressive downsampling. The reconstructed output is aligned with the backbone’s spatial stride, prioritizing structural coherence over pixel-level fidelity. 

To compute the SR reconstruction loss under mismatched spatial resolutions, the input image $X$ is downsampled to match the spatial dimensions of the SR output:
\begin{small}
\begin{equation}
\mathcal{L}_{SR} = \left\| S - \mathcal{D}(X) \right\|_1
\label{eq:SR}
\end{equation}
\end{small}
where $S$ denotes the SR output and $\mathcal{D}(\cdot)$ represents a deterministic downsampling operator applied to the input image $X$ to ensure spatial resolution consistency.

During optimization, gradients from $\mathcal{L}_{SR}$ are propagated into the backbone, reshaping the feature embedding space toward spatially discriminative representations. The auxiliary reconstruction pathway is discarded after training. Consequently, ESM-YOLO+ preserves identical parameter count, computational complexity, and latency as the baseline detector while benefiting from structurally regularized representation learning.

\subsection{Loss Function}

The detection loss $\mathcal{L}$ encompasses three integral components, namely object presence loss $\mathcal{L}_{o}$, object localization loss $\mathcal{L}_{l}$, and object classification loss $\mathcal{L}_{c}$ \cite{Yolov5}, collectively assessing the discrepancy between predictions and ground truth as expressed in \cref{equ:6}.
\begin{small}
\begin{equation}
\mathcal{L}=\lambda_{{o}} \sum_{a=0}^2 \alpha^a_o \mathcal{L}_{{o}} + \lambda_{{l}} \sum_{a=0}^2 \alpha^a_l \mathcal{L}_{{l}} + \lambda_{{c}} \sum_{a=0}^2 \alpha^a_c \mathcal{L}_{{c}}
\label{equ:6}
\end{equation}
\end{small}
where $a$ denotes the index of the output layer within the head; $\alpha^a_o, \alpha^a_l$, and $\alpha^a_c$ are the respective weights for object presence, localization, and classification losses for each layer; and $\lambda_{{o}}, \lambda_{{l}},$ and $\lambda_{{c}}$ serve as global coefficients.

The overall training objective of ESM-YOLO+ consists of the standard detection loss and the auxiliary SR reconstruction loss, which jointly optimize detection accuracy and representation quality:
\begin{small}
\begin{equation}
\mathcal{L}_{total}=c_1 \mathcal{L}+c_2 \mathcal{L}_{SR}
\label{eq:loss}
\end{equation}
\end{small}
where $\mathcal{L}$ denotes the detection loss, $\mathcal{L}_{SR}$ denotes the SR guidance loss, and $c_1$ and $c_2$ are balancing coefficients controlling the importance of the two training objectives.


\section{Result}\label{ch:Result}

\subsection{Datasets}

\textbf{VEDAl dataset}~\cite{Razakarivony2016Vehicle} is designed for vehicle detection in aerial imagery, including complex backgrounds, e.g., grasslands, highways, mountains, and urban landscapes. The dataset consists of 1246 pairs of RGB and infrared images with resolutions of 1024$\times$1024 and 512$\times$512 pixels. In our experiment, we used the 1024$\times$1024 version.

\textbf{DroneVehicle dataset}~\cite{9759286} is currently the largest benchmark for cross-modal vehicle small-target detection. The dataset consists of 56,878 images, namely 28,439 pairs of visible-infrared images. Five types of vehicles are involved, including cars, buses, trucks, vans, and lorries. Among them, the training set, validation set, and test set consist of 17,990, 1469, and 8980 images. The image size is 840$\times$712 pixels. 

\begin{figure}[pos=t]
\centering 
\includegraphics[width=0.8\linewidth]{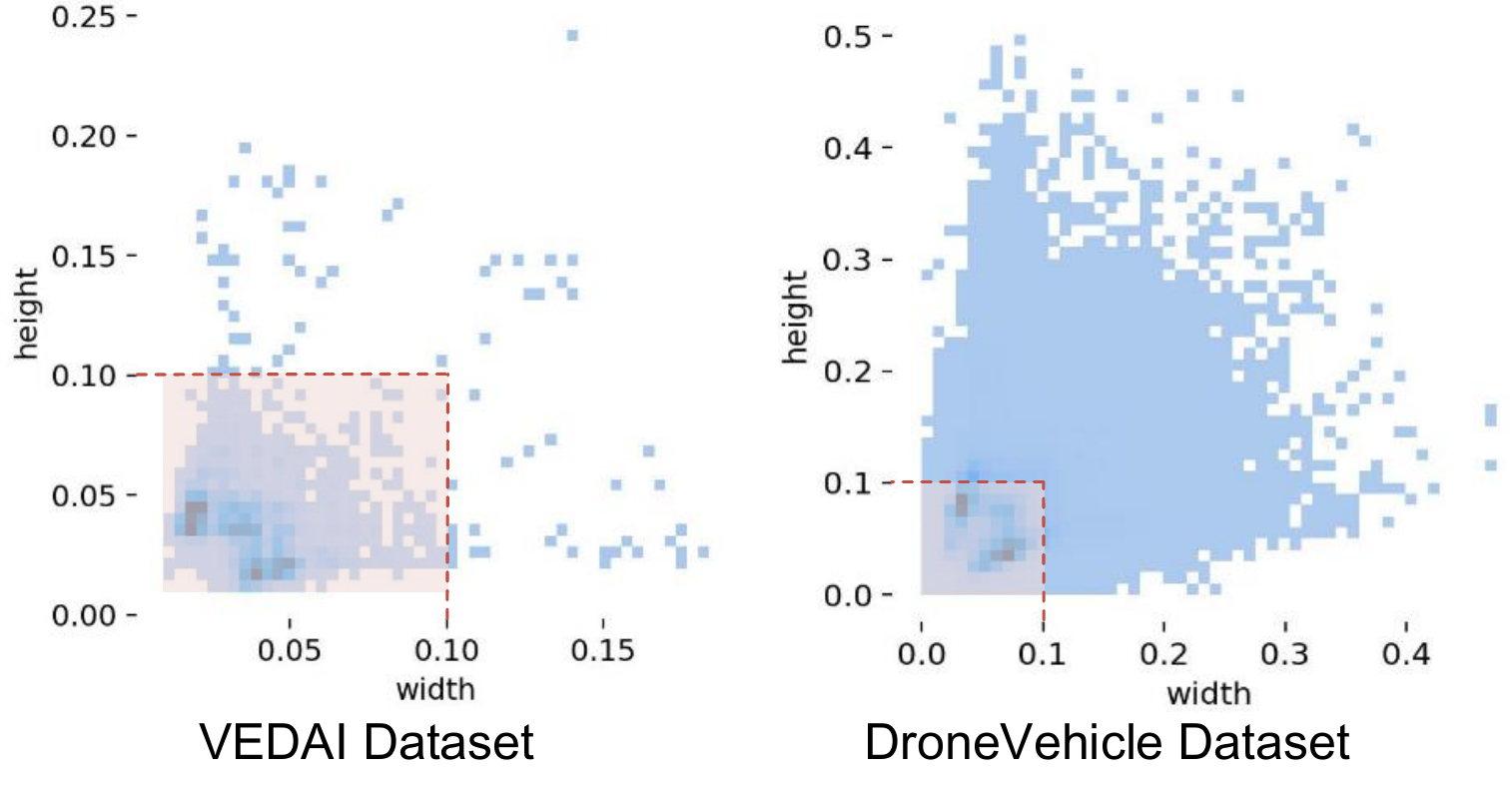}
\caption{Distribution of target sizes in the datasets.} \label{fig:smalltarget}

\end{figure}

As shown in \cref{fig:smalltarget}, the majority of targets in the VEDAI dataset~\cite{Razakarivony2016Vehicle} meet the definition of small targets (with a size ratio below 0.1)~\cite{small}.

\subsection{Training details}
To evaluate the effectiveness of the ESM-YOLO+ models, we conduct comprehensive comparisons and ablation experiments on a machine equipped with an AMD 7945 CPU and an NVIDIA RTX 4060 GPU. The models are trained using the standard stochastic gradient descent (SGD) optimizer~\cite{bottou2010large} with a momentum of 0.937, a weight decay of 0.0005, and Nesterov accelerated gradients. The choice of momentum and learning rate is inspired by the detailed experiments in \cite{standards5010009}, which demonstrate that a learning rate of 0.01 combined with a momentum of 0.937 achieves the highest mAP accuracy. Based on these findings, we use the same values to ensure optimal model performance. The weight decay is set to 0.0005 to prevent overfitting while maintaining model generalization. The training process spans 300 epochs with a batch size of 2, primarily constrained by GPU memory capacity.

For the loss function, the parameters are configured as $\alpha_o^a=1.0$, $\alpha_l^a=0.05$, $\alpha_c^a=0.5$ for $a=0, 1, 2$, and $\lambda_o=\lambda_l=\lambda_c=1.0$. These values are determined through empirical tuning and prior knowledge from related work. Specifically, $\alpha_o^a=1.0$ emphasizes the importance of object presence, while $\alpha_l^a=0.05$ and $\alpha_c^a=0.5$ balance the contributions of object localization and classification losses. The parameters $p^{RGB}$ and $p^{IR}$ in \cref{eq:1} are set to 0.5, reflecting an equal weighting of RGB and infrared modalities in the multimodal fusion process. This choice is based on the assumption that both modalities provide complementary information, and preliminary experiments confirmed that equal weighting achieves a good balance between accuracy and robustness.

In our experiments, we use 1024$\times$1024-resolution images for training and 512$\times$512-resolution images for testing. This choice aim to balance detection accuracy and computational efficiency. High-resolution training images enable the model to capture finer details of small targets, which is essential for improving detection performance. 

For the DroneVehicle dataset, however, we do not perform any resolution adjustment, as the dataset provides only a single resolution of 840$\times$712 pixels. Using the original resolution ensures that the experimental results are consistent with the dataset's characteristics and avoids introducing artifacts from resizing operations.

\subsection{Assessment Indicators}
The accuracy assessment measures the agreements and differences between the detection result and the reference. Recall, precision, and mean Average Precision (mAP) are used as accuracy metrics to evaluate method performance. The calculations of the precision Pr and recall Re metrics are defined as
\begin{small}
\begin{equation}
\text { Pr }=\frac{\mathrm{TP}}{\mathrm{TP}+\mathrm{FP}}, \text { Re }=\frac{\mathrm{TP}}{\mathrm{TP}+\mathrm{FN}}
\end{equation}
\end{small}
where the true positive (TP) and true negative (TN) denote correct prediction, and the false positive (FP) and false negative (FN) denote incorrect outcome. The precision and recall are correlated with the commission and omission errors, respectively. The mAP is a comprehensive indicator obtained by averaging AP values, calculated using an integral method that computes the area under the precision–recall curve for all categories. Hence, the mAP can be calculated by
\begin{small}
\begin{equation}
\text{mAP}=\frac{\text{AP}}{N}=\frac{\int_0^1 \text{Pr}(\text{Re}) d\text{Re}}{N}
\end{equation}
\end{small}
where $N$ is the number of categories. Model complexity and computational cost are quantified using metrics of Giga Floating-point Operations Per Second (GFLOPs) and parameter size.

\subsection{Results Comparisons}
\subsubsection{Comparative Analysis of ESM-YOLO+}

\begin{figure}[pos=h]
\centering 
\includegraphics[width=1\linewidth]{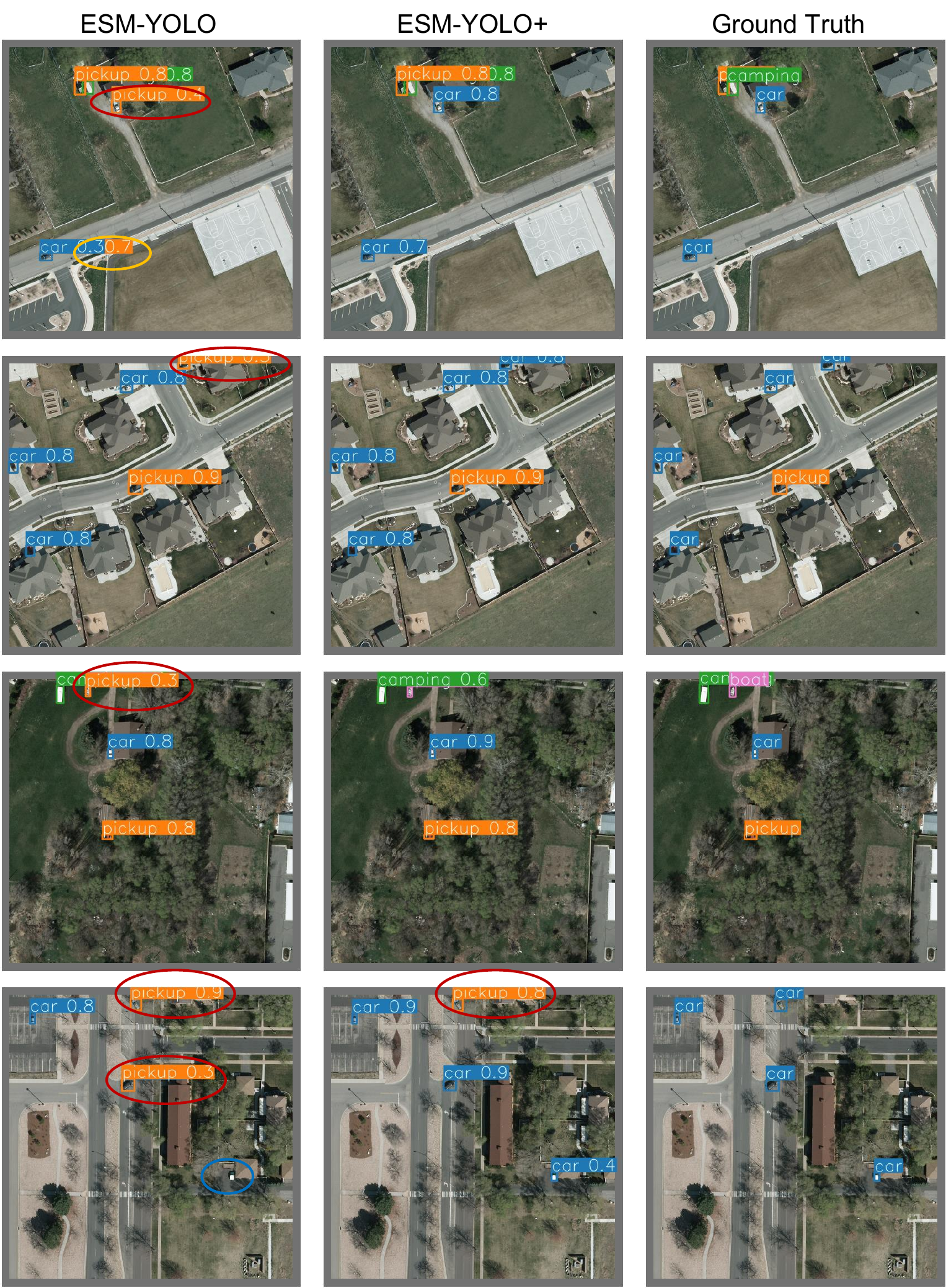} 
\caption{Visual comparison between the ESM-YOLO Model and the ESM-YOLO+ Model. The red cycles represent the false alarms, the yellow ones denote the FP detection results, and the blue ones are FN detection results.} \label{fig:ESMYOLOvsESMYOLO+}
\end{figure}

To evaluate the usability and effectiveness of the ESM-YOLO+ model for detecting small targets in multimodal remote sensing application scenarios, a comparative study was conducted with a variety of algorithms under the same conditions. Detailed experimental comparison results are shown in the \cref{tab:1} and \cref{fig:ESMYOLOvsESMYOLO+}.

\begin{table*}[t]
  \caption{Comparative Experiments of the ESM-YOLO+ model on the VEDAI dataset.}
  \label{tab:1}
  \centering
  \scalebox{0.95}{  
  \begin{tabular}{l|llllllll|l}
    \toprule
    Method & Car & Pickup & Camping & Truck & Other & Tractor & Boat & Van & $mAP_{50}$ ↑ \\
    \midrule
    YOLOv3 \cite{Yolov3} & 84.57 & 72.68 & 67.13 & 61.96 & 43.04 & 65.24 & 37.10 & 58.29 & 61.26 \\
    YOLOv4 \cite{Yolov4} & 85.46 & 72.84 & 72.38 & 62.82 & 48.94 & 68.99 & 34.28 & 54.66 & 62.55 \\
    YOLOv5s \cite{Yolov5} & 80.81 & 68.48 & 69.06 & 54.71 & 46.79 & 64.29 & 24.25 & 45.96 & 56.79 \\
    YOLOv5m \cite{Yolov5} & 82.53 & 72.32 & 68.41 & 59.25 & 46.20 & 66.23 & 33.51 & 57.11 & 60.69 \\
    YOLOv5l \cite{Yolov5} & 82.83 & 72.32 & 69.92 & 63.94 & 48.48 & 63.07 & 40.12 & 56.46 & 62.16 \\
    YOLOv5x \cite{Yolov5} & 84.33 & 72.95 & 70.09 & 61.15 & 49.94 & 67.35 & 38.71 & 56.65 & 62.65 \\
    SuperYOLO \cite{SuperYOLO} & 91.13 & 85.66 & 79.30 & 70.18 & 57.33 & 80.41 & 60.24 & 76.50 & 75.09 \\
    ESM-YOLO\cite{esm} & 90.80 & 87.79 & 83.31 & 83.83 & 69.48 & 78.78 & \textbf{85.23} & 80.11 & 82.42 \\
    \textbf{ESM-YOLO+} & \textbf{93.64} & \textbf{88.76} & \textbf{85.66} & \textbf{86.65} & \textbf{74.80} & \textbf{91.39} & 71.57 & \textbf{85.20} & \textbf{84.71} \\
    \bottomrule
  \end{tabular}
  }
\end{table*}


\begin{figure}[pos=h]
\centering 
\includegraphics[width=1\linewidth]{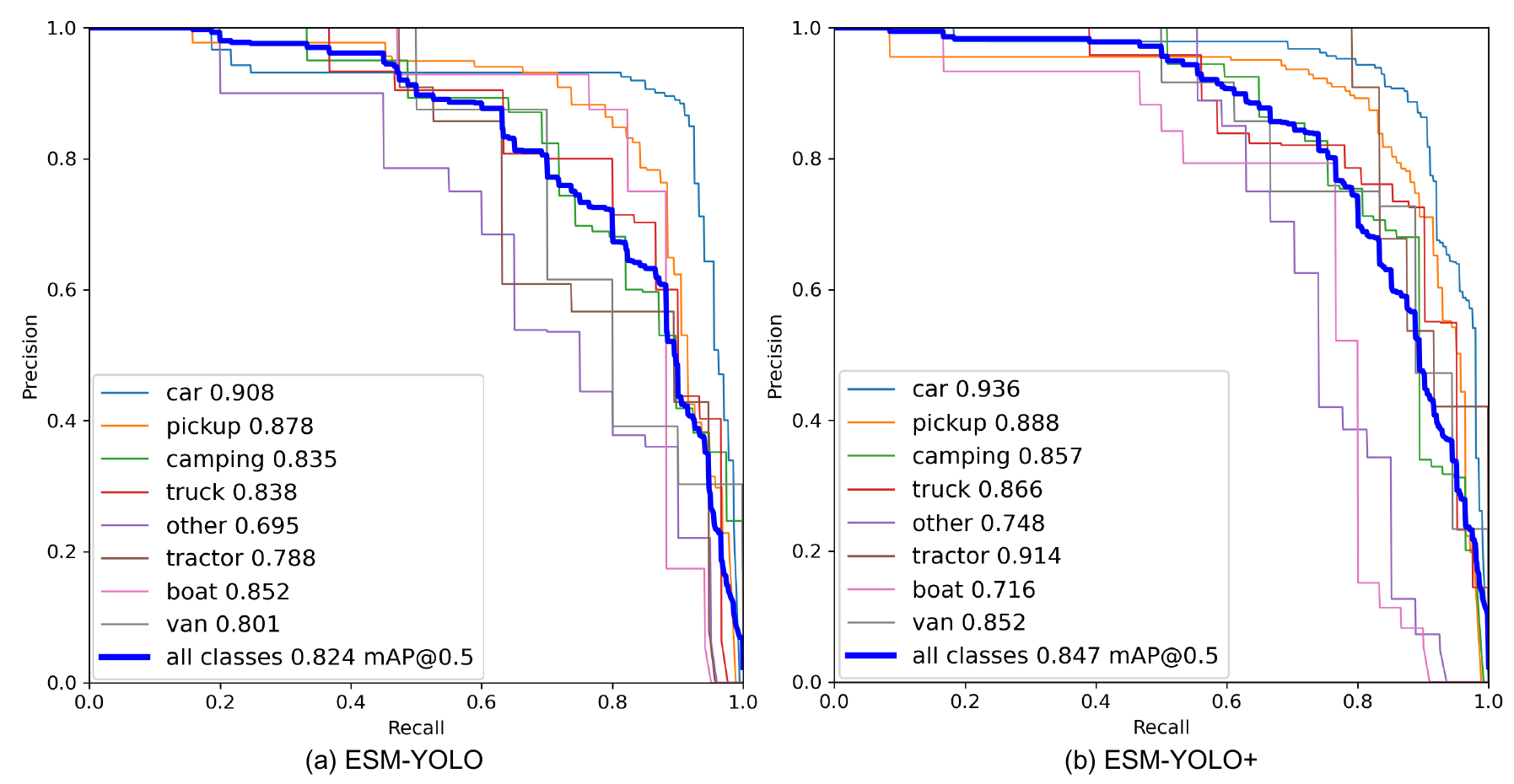}
\caption{Comparison of P-R Curve for ESM-YOLO (a) and ESM-YOLO+ (b).} \label{fig:2}
\end{figure}

The P-R curve reflects the model's ability to distinguish between positive and negative samples. The P-R curve of the ESM-YOLO+ model is closer to the upper-left corner, indicating better balance between accuracy and recall and overall performance. As shown in the \cref{fig:2}.

\begin{figure}[pos=h]
\centering 
\includegraphics[width=1\linewidth]{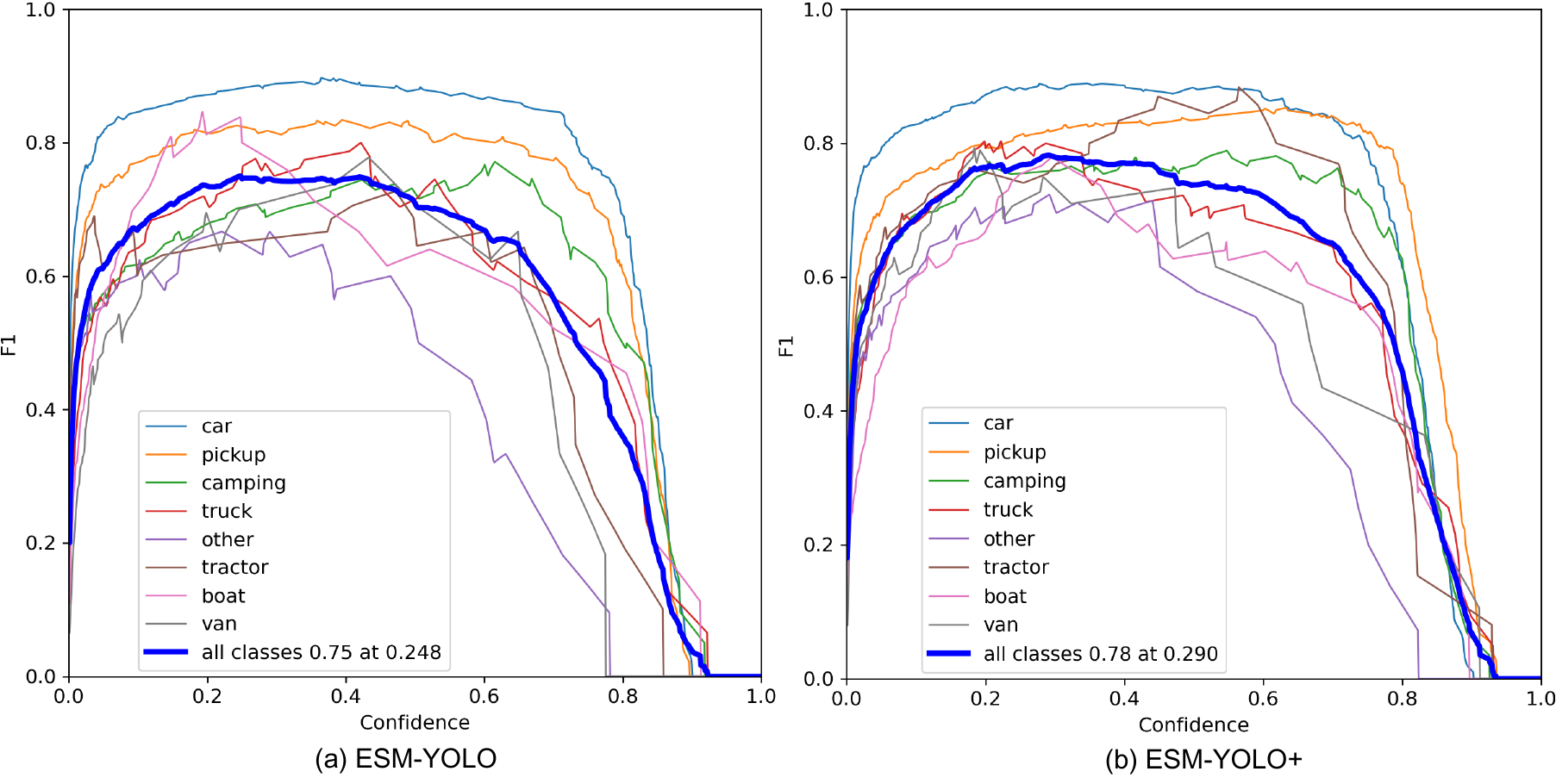}
\caption{Comparison of F1 curves for ESM-YOLO (a) and ESM-YOLO+ (b).} \label{fig:F1}
\end{figure}

The F1 curve can show how the model's F1 score changes across different classification thresholds. The F1 score is the harmonic mean of precision and recall and is used to evaluate the performance of a classification model. The F1 score of the ESM-YOLO+ model is higher than that of the ESM-YOLO model, indicating that the ESM-YOLO+ model has a better balance between accuracy and recall, and has better performance. As shown in the \cref{fig:F1}.

\subsubsection{Influence of Multimodalities}
To validate the substantial enhancement in recognition accuracy for small target detection achieved by fused multimodal data over unimodal data after network training, a comparative analysis is conducted under identical conditions. This analysis encompasses the detection outcomes of unimodal visible, unimodal infrared, and multimodal fusion-based small target recognition. Specifically, the mean average precision (mAP) for the unimodal visible model is 75.06\%, while that of the unimodal infrared model stands at 70.83\%. Notably, the multimodal ESM-YOLO+ model demonstrates a significant improvement in small-target detection accuracy. The comprehensive experimental results are presented in \cref{tab:3}.



\begin{table*}[t]
  \caption{Comparison of single modality and multimodalities (ESM-YOLO+) on the VEDAI dataset.
  }
  \label{tab:3}
  \centering
  \scalebox{1}{
  \begin{tabular}{@{}l|llllllll|l@{}}
    \toprule
     & Car & Pickup & Camping & Truck & Other & Tractor & Boat & Van & $mAP_{50}$ ↑\\
    \midrule
    Single-Mode (RGB) & 89.31 & 87.31 & 69.71 & 83.34 & 61.04 & 80.08 & 54.17 & 75.52 & 75.06 \\
    Single-Mode (IR) & 87.84 & 86.00 & 78.54 & 72.09 & 45.44 & 59.95 & 55.02 & 81.77 & 70.83 \\
    \textbf{MultiModal (RGB+IR)} & \textbf{93.64} & \textbf{88.76} & \textbf{85.66} & \textbf{86.65} & \textbf{74.80} & \textbf{91.39} & \textbf{71.57} & \textbf{85.20} & \textbf{84.71} \\
  \bottomrule
  \end{tabular}
  
  }
\end{table*}

\begin{figure}[pos=h]
    \centering
    \includegraphics[width=1\linewidth]{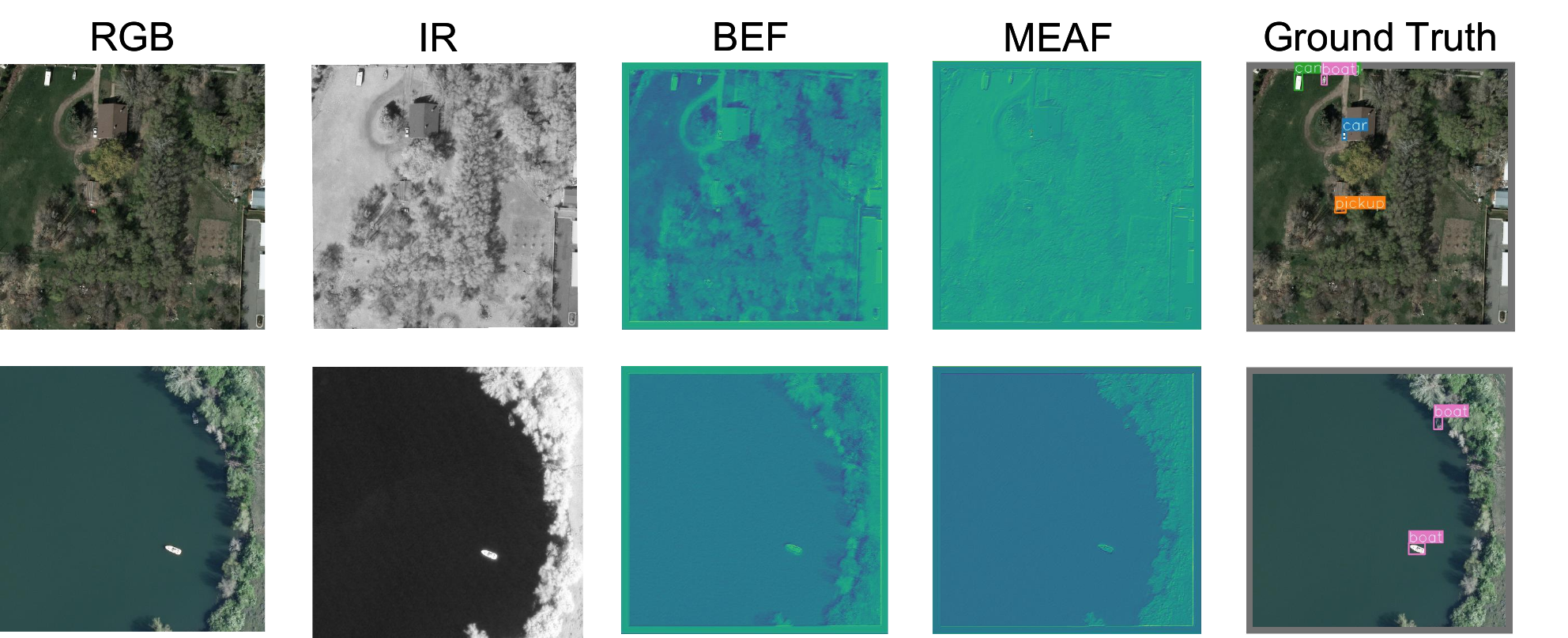}
    \caption{Visualization of BEF fusion and MEAF fusion results.}
    \label{fig:5}
   
\end{figure}

To evaluate the fusion quality of the BEF fusion method and the MEAF fusion method. We did a comparison experiment under the same conditions. As shown in the \cref{fig:5}. MEAF fusion method can better eliminate the interference of shadow and complex background, so as to generate a higher quality fusion image.

ESM-YOLO+ achieves a strong balance between accuracy and efficiency on the VEDAI~\cite{Razakarivony2016Vehicle} dataset. As shown in \cref{tab:VEDAItransformer}, our model achieves a competitive mAP of 84.7\% with only 5.1M parameters, demonstrating superior efficiency compared to state-of-the-art Transformer-based methods. Transformer-based methods such as CFT~\cite{wang2024cross}, Cross-Channel-ViT~\cite{10647683}, and ICAFusion~\cite{shen2024icafusion} require heavy computational resources, while hybrid Transformer + CNN models like MOD-YOLO~\cite{shao2024mod} and MOD-YOLO-Tiny~\cite{shao2024mod} achieve substantially lower accuracy, demonstrating that these approaches are either inefficient or less effective compared to ESM-YOLO+.
\begin{table}[htbp]
  \caption{Comparative Experiments of the ESM-YOLO+ model on the VEDAI dataset.}
  \label{tab:VEDAItransformer}
  \centering
  \resizebox{\columnwidth}{!}{ 
  \begin{tabular}{l c l l}
    \toprule
    Method & Backbone &  $mAP_{50}$ ↑ & Params(M) ↓ \\
    \midrule
    CFT~\cite{wang2024cross} & Transformer  & \textbf{85.3} & 196.9 \\
    ICAFusion~\cite{shen2024icafusion} & Transformer & 76.6 & 120.2 \\
    Cross-Channel-ViT~\cite{10647683} & Transformer & 78.5 & - \\
    \midrule
    MOD-YOLO~\cite{shao2024mod} & Transformer + CNN & 59.3 & 24.9 \\
    MOD-YOLO-Tiny~\cite{shao2024mod} & Transformer + CNN & 58.4 & 16.0 \\
    \midrule
    \textbf{ESM-YOLO+} & CNN &  84.7 & \textbf{5.1} \\
    \bottomrule
  \end{tabular}}
\end{table}

The experimental results in Table~\ref{tab:VEDAIsmalltarget} demonstrate that ESM-YOLO+ achieves state-of-the-art performance in small target detection, with a significant $\text{mAP}_{50}$ improvement of 6.6\% over the second-best method (ACDF-YOLO). The consistent superiority across 6 out of 8 categories validates the robustness of our approach.

\begin{table*}[htbp]
  \caption{Comparison of Detection Accuracy Performance between Specialized Algorithms for Small Target Detection in Remote Sensing MultiModal Fusion and  Our Method
  }
  \label{tab:VEDAIsmalltarget}
  \centering
  \scalebox{0.85}{
  \begin{tabular}{@{}l|llllllll|l@{}}
    \toprule
   Method  & Car & Pickup & Camping & Truck & Other & Tractor & Boat & Van & $mAP_{50}$ ↑\\
    \midrule
   MIR-YOLO~\cite{zhong2025mir}  & 88.8 & 77.2 & 70.8 & 68.7 & 71.0 & 72.8 & 60.8 & 47.8 & 69.7 \\
   ACDF-YOLO~\cite{fei2024acdf} & 89.8 & 87.9 & 78.5 & 75.3 & 67.3 & 76.3 & 73.7 & 75.9 & 78.1 \\
   MMFDet~\cite{zhao2025differential} & 88.3 & 78.5 & 81.6 & 59.8 & 63.5 & 86.2 & \textbf{76.0} & \textbf{88.3} & 77.9 \\
   FFCA-YOLO~\cite{10423050} & 89.6 & 85.7 & 78.7 & 85.7 & 48.6 & 81.8 & 61.5 & 67.0 & 74.8 \\
   L-FFCA-YOLO ~\cite{10423050} & 91.3 & 85.5 & 72.8 & 79.7 & 47.3 & 79.0 & 56.1 & 73.9 & 73.3 \\
   \textbf{ESM-YOLO+} & \textbf{93.6} & \textbf{88.8} & \textbf{85.7} & \textbf{86.7} & \textbf{74.8} & \textbf{91.4} & 71.6 & 85.2 & \textbf{84.7} \\
  \bottomrule
  \end{tabular}
  
  }
\end{table*}

In order to verify that our model still performs well on other datasets, we conducted a comparative experiment on the DroneVehicle~\cite{9759286} dataset. The detailed experimental results are presented in \cref{tab:DroneVehicle}.
\begin{table*}[t]
  \caption{Comparative Experiments of the ESM-YOLO+ model on the DroneVehicle dataset.
  }
  \label{tab:DroneVehicle}
  \centering
  \scalebox{0.8}{
  \begin{tabular}{l|c|lllll|lll}
    \toprule
    Method & Architecture & Car & Truck & Bus & Van & Freight Car  & $mAP_{50}$ ↑ & Params(M) ↓  & GFLOPs(G) ↓\\
 
    \midrule
      
         UA-CMDet~\cite{9759286}&  \multirow{4}{*}{CNN} & 87.5 & 60.7 & 87.1 & 37.9 & 46.8  & 64.0 & 234.0 & - \\
         MKD~\cite{huang2023multimodal}& & 93.5 & 62.5  & 91.9 & 44.5 &  52.7 & 69.0 & 242.0 & -\\ 
         Oriented R-CNN~\cite{9710901}& & 89.9 & 56.6 & 89.6& 46.9 & 54.4 & 67.5 & 41.1 & 107.3 \\
        TSFADet~\cite{yuan2022translation}&  & 89.9 &67.9 & 89.9 & 54.0 & 63.7 & 73.1 & 104.7 & 109.8\\
 \midrule
  C\textsuperscript{2}Former-S\textsuperscript{2}ANet~\cite{yuan2024c}&  \multirow{3}{*}{Transformer} & 90.2 & 68.3 & 89.8 & 58.5 & \textbf{64.4}  & 74.2 & 132.5 & 100.9\\
  ViT-B+RVSA~\cite{9956816}& & 89.7 & 52.3 & 88.0 & 44.4 & 51.0 & 65.1 & 114.4 & 416.6\\
           VIP-Det~\cite{chen2024drone}& &90.4  & \textbf{78.5} & 89.8  & 57.5 &  61.4 &  \textbf{75.5} &  70.1 &  - \\
  \midrule
          \textbf{ESM-YOLO+}& CNN & \textbf{97.1} &67.9& \textbf{94.4} & \textbf{64.5} & 45.9 &74.0& \textbf{5.1} & \textbf{20.8}\\
  \bottomrule
  \end{tabular}
  }
  
\end{table*}
As shown in \cref{tab:DroneVehicle}. The parameter count of ESM-YOLO+ is only 5.1M, the lowest among all compared methods.   The number of parameters reflects the scale of parameters that the model needs to store and update.   This indicates that ESM-YOLO+ adopts a relatively concise model architecture, effectively reducing storage requirements and computational costs while maintaining detection performance.   In contrast, CNN-based methods such as UA-CMDet~\cite{9759286} (234.0M) and MKD~\cite{huang2023multimodal} (242.0M), as well as Transformer-based methods like ViT-B+RVSA~\cite{9956816} (114.4M) and $C\textsuperscript{2}Former-S\textsuperscript{2}ANet$~\cite{yuan2024c} (132.5M), have significantly higher parameter counts than ESM-YOLO+.   This is due to their more complex network structures or additional feature extraction modules, leading to increased parameter sizes.

The GFLOPs of ESM-YOLO+ is 20.8G, also among the lowest in all compared methods.   GFLOPs reflect the computational load required during the model's inference process.   This demonstrates that ESM-YOLO+ requires relatively low computational effort during inference.   CNN-based methods such as Oriented R-CNN~\cite{9710901} (107.3G) and TSFADet~\cite{yuan2022translation} (109.8G), as well as Transformer-based methods like ViT-B+RVSA~\cite{9956816} (416.6G), have significantly higher GFLOPs than ESM-YOLO+.   This indicates that these methods demand more computational resources during inference, which limit their real-time performance and deployment feasibility in practical applications.
 

  


\begin{figure}[pos=h]
\centering 
\includegraphics[width=1\linewidth]{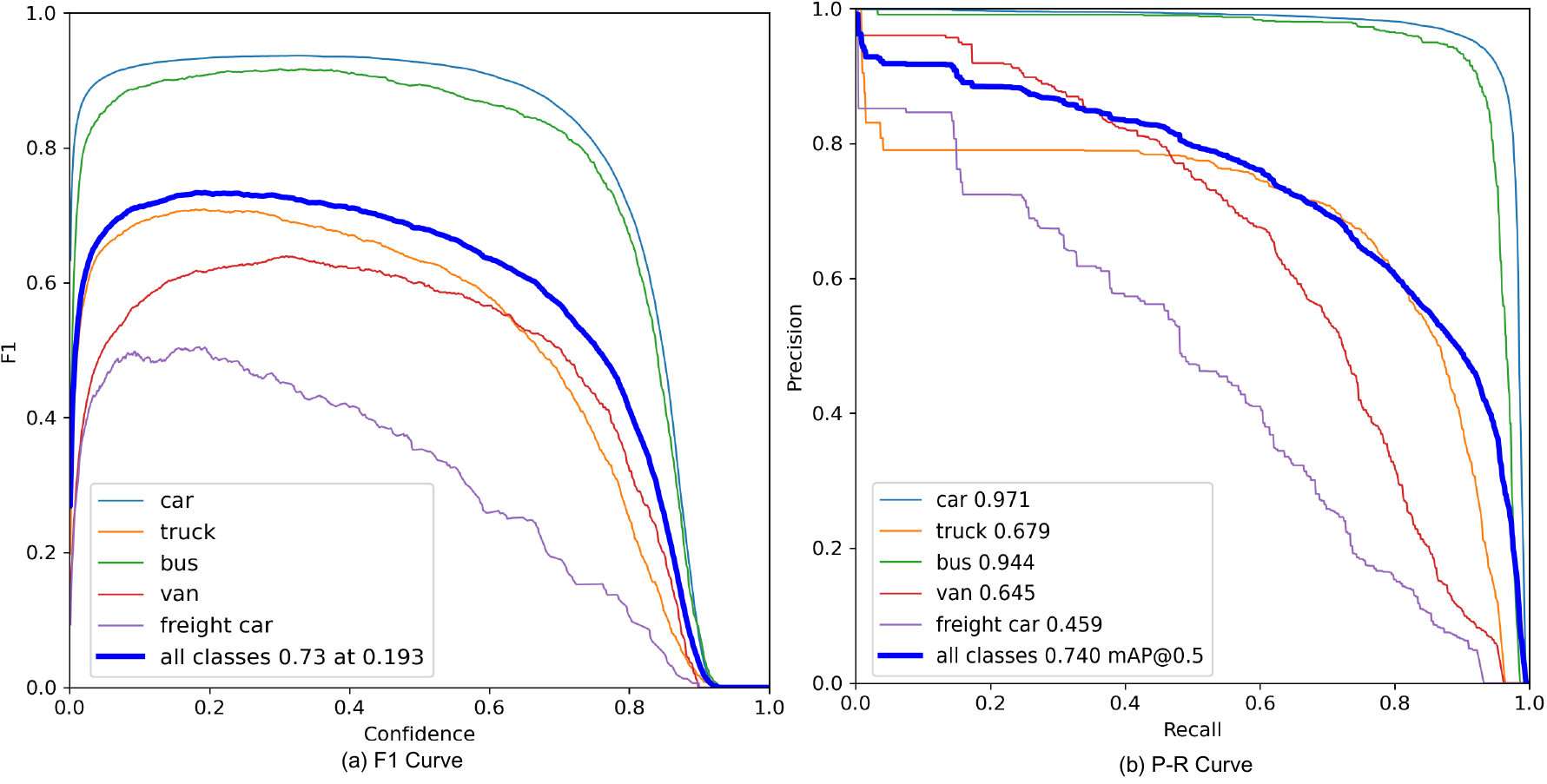}
\caption{The F1 curve and P-R curve obtained by the ESM-YOLO+ model validated on the DroneVehicle dataset.} \label{fig:dronevehicle}

\end{figure}

\begin{figure}[pos=h]
\centering 
\includegraphics[width=1\linewidth]{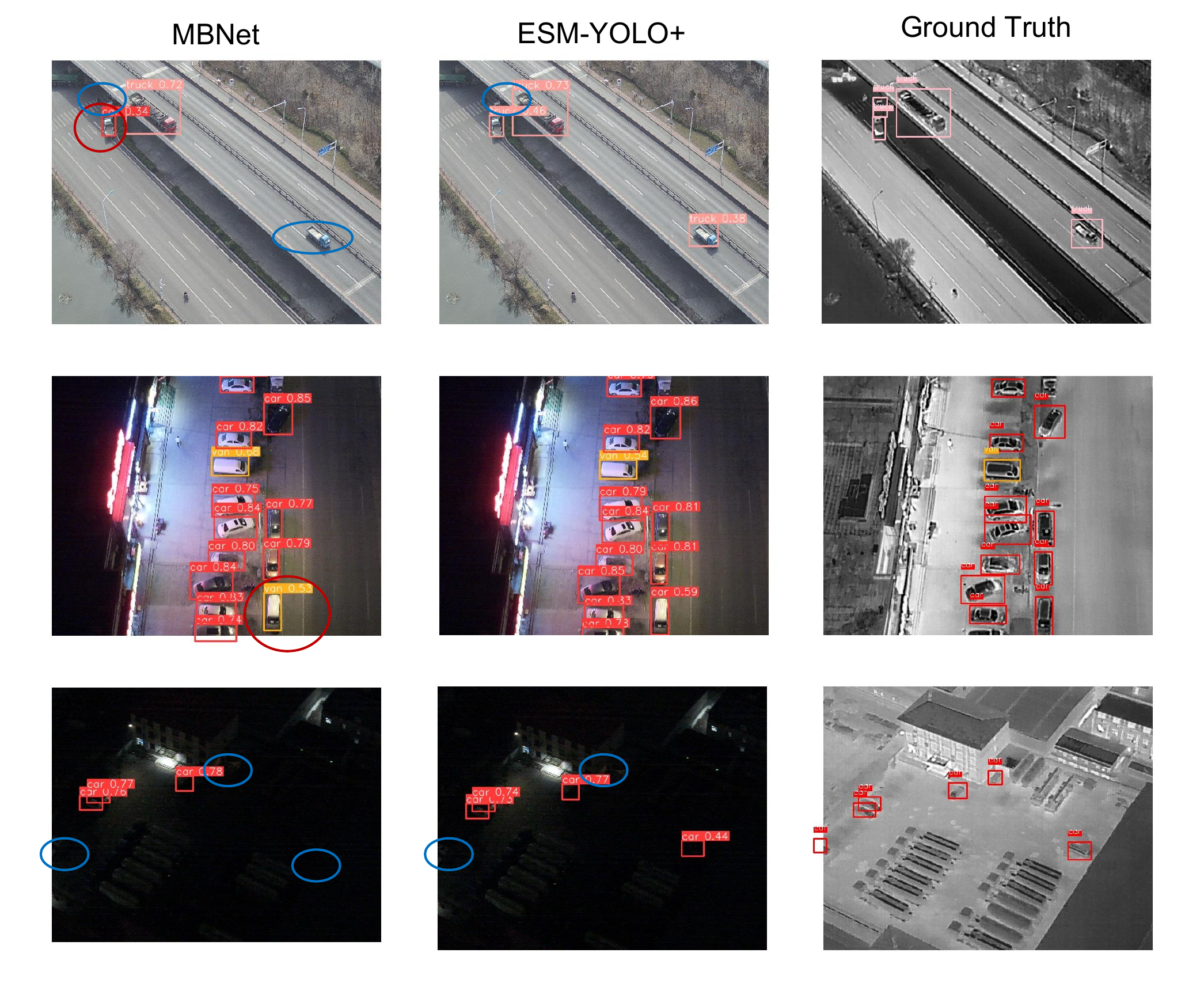}
\caption{The visualization shows the comparison results of ESM-YOLO+ and other algorithms verified on the DroneVehicle dataset~\cite{9759286}. The red cycles represent the false alarms, the yellow ones denote the FP detection results, and the blue ones are FN detection results.} \label{fig:dronevehicleVS}

\end{figure}

\cref{fig:dronevehicle} shows the F1 and P-R curves of the ESM-YOLO+ model when evaluated on the DroneVehicle~\cite{9759286} dataset. \cref{fig:dronevehicleVS} visually shows the detection performance advantage of ESM-YOLO+ model when it is evaluated on DroneVehicle dataset.


\begin{table}[htbp]
  \caption{Ablation experiments on MEAF Module. EB: ESM-YOLO Backbone; EB (-SR): ESM-YOLO Backbone (SR removed during inference)}
  \label{tab:MEAF}
  \centering
  \resizebox{\columnwidth}{!}{
  \begin{tabular}{l|l l ll |l|l}
    \toprule[1pt]
 No.  & EB & EB (-SR)  & BEF & MEAF 
   & $mAP_{50}$ ↑& Params(M) ↓ \\
    \hline
1 & \checkmark & - &  - &  -   & 72.68 & 80.16\\
2 & \checkmark & - &  \checkmark &  -   & 82.42 & 80.18\\
3 & - & \checkmark &  - &   \checkmark   & \textbf{84.71} & 5.14\\    
  \bottomrule[1pt]
  \end{tabular}
  }
\end{table}
\subsection{Ablation Study}

\textbf{Impact of Mask-Enhanced Attention Fusion}
Further augmentation is observed with the integration of the MEAF module, which elevates the mAP to 84.71\%. This increment verifies that MEAF can effectively improve the quality of visible and infrared modal fusion images and can effectively improve the detection accuracy of small targets. As shown in the \cref{tab:MEAF}.

\section{Discussion}
\label{ch:discussion} 
The core value of the lightweight visible-infrared fusion network ESM-YOLO+ proposed in this study is to solve the small target detection in remote sensing images: weak texture of small targets, easy interference from complex backgrounds, alignment deviation, and scale heterogeneity in cross-modal fusion. It also balances model performance and practicality in real deployment, providing a new technical idea for small target detection in complex remote sensing scenarios.

The design of the Mask-Enhanced Attention Fusion (MEAF) module breaks the limitation of traditional fusion methods that only focus on feature superposition and ignore pixel-level alignment and feature enhancement. By combining learnable spatial masks and spatial attention mechanisms, it achieves accurate alignment of RGB and infrared features, which not only strengthens the feature expression of small targets but also effectively reduces fusion distortion caused by cross-modal differences. The advantage of this fusion method is that it can complement two modal information without adding complex alignment networks, allowing the model to capture weak features of small targets more clearly, thereby improving detection accuracy in complex backgrounds. This is one of the key reasons why ESM-YOLO+ achieves performance improvement on remote sensing datasets with high proportion of small targets.

The Structural Representation (SR) enhancement strategy introduced during training builds an efficient auxiliary supervision mechanism. Its core significance is to achieve a balance of "enhancement during training, no burden during inference". During training, SR guidance can prompt the model to focus on the fine-grained spatial structure of small targets and improve feature discriminability. In the inference stage, no additional super-resolution operations are needed, avoiding the increase of inference cost. This design accurately meets the actual needs of real-time remote sensing detection: ensuring detection accuracy while meeting the speed requirements during deployment.

The performance improvement and lightweight model optimization reflected in the experimental results further verify the rationality and practicality of the design idea of this study. Compared with the basic model ESM-YOLO, ESM-YOLO+ achieves a 2.29\% increase in mAP on the VEDAI dataset, while reducing 93.6\% parameters and 68.0\% GFLOPs. This dual optimization of "performance improvement + complexity reduction" breaks the dilemma in traditional remote sensing target detection models that "high precision is necessarily accompanied by high complexity". This means that ESM-YOLO+ can adapt to more resource-constrained scenarios, such as real-time UAV detection and rapid processing of satellite remote sensing images, greatly expanding the practical application scope of the model and providing a feasible solution for the engineering implementation of remote sensing small target detection technology.

It should be noted that the experiments in this study are mainly based on two public datasets, VEDAI and DroneVehicle. Although these two datasets cover different remote sensing scenarios and target types, the detection performance of the model may be affected in more complex extreme environments (such as strong light, heavy fog, heavy rain, etc.), which is a direction that needs further optimization in future research. In addition, how to further improve the fusion efficiency of the MEAF module and combine the SR guidance strategy with other training optimization methods to further tap the performance potential of the model is also worthy of in-depth exploration in the follow-up.

\section{Conclusion}\label{ch:Conclusion}

To address weak texture, complex backgrounds, and cross-modal misalignment in remote sensing small-object detection, we propose ESM-YOLO+, a lightweight visible–infrared fusion framework that strengthens structurally consistent cross-modal representations and preserves fine-grained spatial information during training, enabling improved small-target discriminability under resource-constrained deployment. Experiments on the VEDAI and DroneVehicle datasets show that ESM-YOLO+ improves small-object detection accuracy over the baseline while maintaining inference-time efficiency, ensuring real-time applicability. These results confirm that the proposed ESM-YOLO+ effectively balances representational richness and computational constraints, providing a practical approach for real-time resource-limited remote sensing applications.

\section*{Declaration of competing interest} 
The authors declare that they have no conflicts of interest in this work.


\bibliographystyle{cas-model2-names}
\bibliography{main}


\end{document}